\documentclass[conference]{IEEEtran}

\usepackage{cite}
\usepackage{amsmath,amssymb,amsfonts}
\usepackage{algorithmic}
\usepackage{graphicx}
\usepackage{textcomp}
\usepackage{xcolor}

\usepackage[caption=false]{subfig}
\usepackage{multirow}
\usepackage{graphicx}

\def\BibTeX{{\rm B\kern-.05em{\sc i\kern-.025em b}\kern-.08em
    T\kern-.1667em\lower.7ex\hbox{E}\kern-.125emX}}
\begin{document}

\title{MT-SNN: Spiking Neural Network that Enables Single-Tasking of Multiple Tasks
}

\author{
\IEEEauthorblockN{Paolo G. Cachi}
\IEEEauthorblockA{\textit{Department of Computer Science}\\
\textit{Virginia Commonwealth University}\\
Richmond, United States\\
pcachi@vcu.edu}
\and
\IEEEauthorblockN{Sebastián Ventura}
\IEEEauthorblockA{\textit{Department of Computer Science} \\
\textit{Universidad de Córdoba}\\
Córdoba, Spain \\
sventura@uco.es}
\and
\IEEEauthorblockN{Krzysztof J.~Cios}
\IEEEauthorblockA{\textit{Department of Computer Science}\\
\textit{Virginia Commonwealth University}\\
Richmond, United States\\
\textit{University of Information, Technology}\\
\textit{and Management}, Rzeszow, Poland\\
kcios@vcu.edu}
}

\maketitle

\begin{abstract}



In this paper we explore capabilities of spiking neural networks in solving multi-task classification problems using the approach of single-tasking of multiple tasks. We designed and implemented a multi-task spiking neural network (MT-SNN) that can learn two or more classification tasks while performing one task at a time. The task to perform is selected by modulating the firing threshold of leaky integrate and fire neurons used in this work. The network is implemented using Intel’s Lava platform for the Loihi2 neuromorphic chip. Tests are performed on dynamic multitask classification for NMNIST data. The results show that MT-SNN effectively learns multiple tasks by modifying its dynamics, namely, the spiking neurons’ firing threshold.

\end{abstract}


\section{Introduction}

Multi-task learning is a machine learning problem in which a model is trained to solve more than one task \cite{Crawshaw2020}. This forces the model to learn more general and robust representations required for performing such multiple tasks. Multi-task learning is thought to be closer to the operation of a biological brain than single-task learning \cite{Crawshaw2020}. Combining learning of multiple tasks, however, encounters problems such as negative transfer, i.e., when different tasks have conflicting needs, such as when increasing performance for one task decreases performance of the other task(s).

Several solutions were proposed to deal with the negative learning problem \cite{Bilen2017, Liu2017, Rebuffi2018, Maninis2019}. The solution of interest here is \cite{Maninis2019}, where the authors solved the multi-task learning problem using an approach called single tasking of multiple tasks. It consists of training a neural network to solve more than one task but doing only one task at a time, which is controlled by an external additional input. They used attention-like mechanisms in combination with adversarial loss for training a feed-forward neural network to learn task-specific features. In other words, by using the attention-like mechanisms the network selects different set of features for learning each task. In this way it creates internal pathways that process different tasks independently mitigating the negative transfer problem.

The above described solution was implemented using classical, non-spiking, neural networks. In this work we use spiking neural networks that are much more energy efficient  when run on a neuromorphic chip/computer; such as Intel’s Loihi2 chip \cite{Davies2021}. A spiking neuron is a dynamic unit that operates through time similar to operation of biological neurons. Complexity of spiking neurons allows for more powerful computations, but their understanding is still in an early stage. Although solutions using spiking neural networks to simple machine learning tasks were reported \cite{Tavanaei2019, Shin2010, Shin1995}, solutions to more complex tasks, such as the multi-task learning problem, have not been done yet.

In this paper, with the aim of better understanding and thus allow for a wider use of spiking neural networks, we present our findings of using them for solving multi-task classification problems. The presented solution is constructed based on the single tasking of multiple task approach. The network behavior, in terms of which classification task to perform, is controlled by modulating the spiking neuron’s firing threshold, which can be seen as a simple implementation of the neuromodulation property of biological neurons \cite{Marder2012}. The MT-SNN architecture, shown in Figure \ref{fig:MultitaskArchitecture}, consists of three blocks. Each block consists of one or more spiking neuron layers connected in a feed-forward fashion. SLAYER, a backpropagation algorithm for spiking neural networks, is used for training the system \cite{Shrestha2018}. The MT-SNN is implemented in Intel’s Lava neuromorphic framework which will allow for its testing directly on the Loihi2 chip once the access to the chip is granted. MT-SNN is tested on the multi-task classification NMNIST data \cite{Garrick2015}.

The specific contributions of this work are as follows.

• We propose a spiking neural network that performs multiple classification tasks based on controlling the neuron’s firing threshold.

• We show that using an additional block for task classification improves performance losses caused by the negative transfer problem inherent in solving multi-task problems.

• We experimentally show that modifying the firing threshold of neurons gives better results than modifying an external input to control MT-SNN operation.

The paper is structured as follows. Section 2 describes the proposed multi-task spiking neural network. Section 3 describes experimental settings and results. The paper ends with conclusions.

\section{Method}

\subsection{Problem Definition}

In the setting of single tasking of a multi-task problem, we consider an input space  $X$, where $X \in \mathbb{R}^n$ and a set of two (or more) classification labels $Y^{(1)}$ and $Y^{(2)}$, where $Y^{(1)} = \{y_1^{(1)}, y_2^{(1)}, ... , y_m^{(1)}\}$ and $Y^{(2)} = \{y_1^{(2)}, y_2^{(2)}, ... , y_p^{(2)}\}$. Then we want to construct a spiking neural network, $F$, with weights $W$ and an internal parameter (in our case the firing threshold) $\varphi$ that learns the transformations: $y_i^{(1)} = F(x_i \mid W, \varphi=\varphi_1)$ and $y_i^{(2)} = F(x_i \mid W,\varphi=\varphi_2)$.  

\subsection{Model} \label{sec:model}

\begin{figure*}
\centering
\includegraphics[clip, width=1.7\columnwidth]{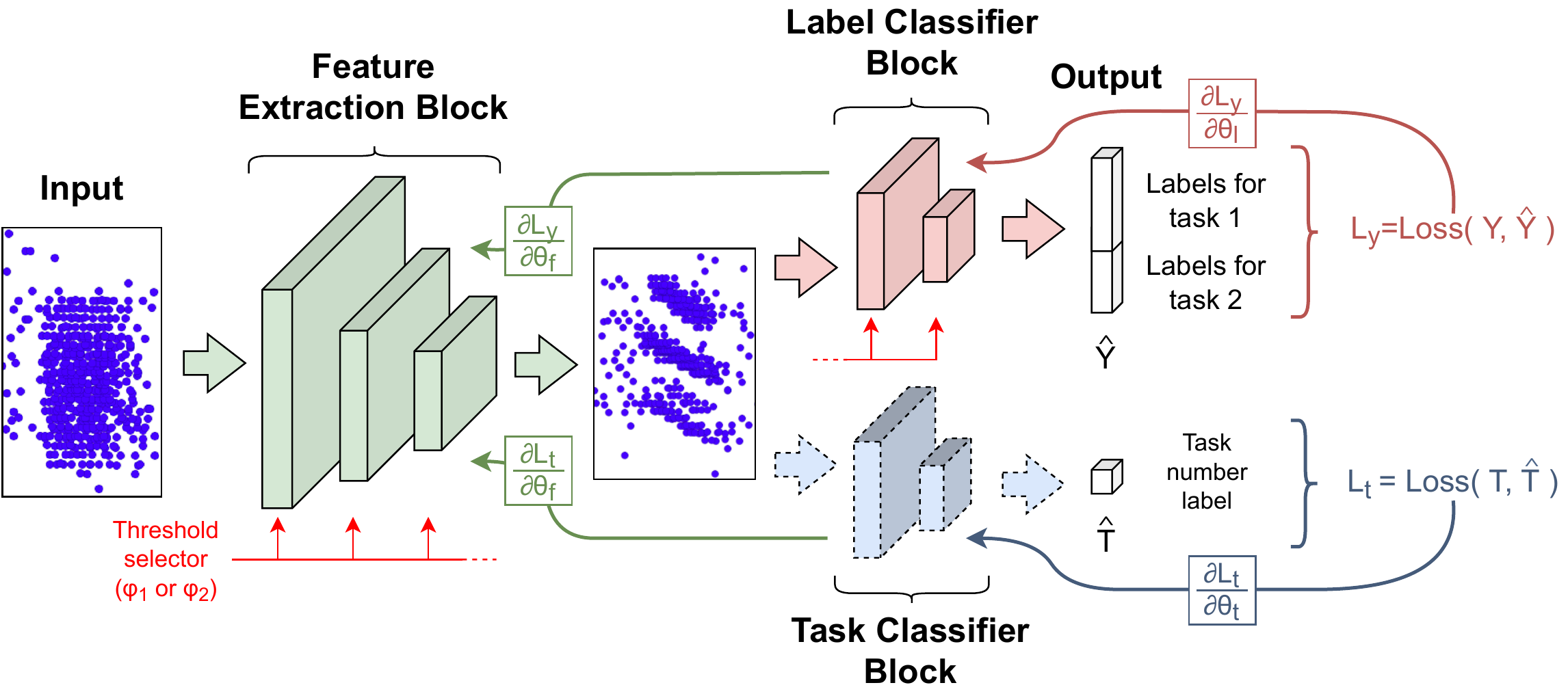}
\caption{MT-SNN architecture. The network consists of three processing blocks connected in a feed-forward fashion: a feature extraction and two classifier blocks. The label classifier outputs the labels for task 1 or task 2 (or more). The task classifier is used as a regularization mechanism to aid the feature extraction block learning a set of independent features for each particular task.} 
\label{fig:MultitaskArchitecture}
\end{figure*}

The MT-SNN architecture is shown in Figure \ref{fig:MultitaskArchitecture}. It consists of three spiking neuron blocks connected in a feed-forward fashion, similar to \cite{Ganin2015}. The spiking input signal is processed by the first block, the feature extraction block, into a latent $p$-dimensional spiking feature vector, which is then used to assign the multi-task labels using a label classifier block. A task classifier block is used for learning the specific task that is being performed. The idea behind this 3-block architecture is to allow the feature extraction block receive feedback from the label classifier block (label classification loss) and also from the task classifier block (task loss) during training. In this way, the task classifier block acts as a regularization block for the feature extraction block. The task classifier block is not used during testing. Note that in contrast to the architecture proposed in \cite{Ganin2015}, MT-SNN does not use a gradient reversal layer before the task classifier block. This is done to enforce the feature extraction block learn independent feature vectors for each classification task.

As the spiking neuron model in MT-SNN, the discrete-time approximation of the integrate and fire neuron model \cite{Kaiser2020} is used. The equations for the membrane potential, $U_i^{(l)}[n]$, and synaptic current, $I_i^{(l)}[n]$, of neuron $i$ in layer $l$ are given by:

\begin{gather}
U_i^{(l)}[n+1] = \alpha U_i^{(l)}[n] + I_i^{(l)}[n] - S_i^{(l)}[n]    \label{eq:MembranePotential}\\
\resizebox{.91\linewidth}{!}{$
    \displaystyle
    I_i^{(l)}[n+1] = \beta I_i^{(l)}[n] + \sum_j W_{ij}^{(l)}S_j^{(l)}[n] + \sum_j V_{ij}^{(l)}S_j^{(l)}[n]
$} \label{eq:SynapticCurrent}
\end{gather}

where $\alpha$ and $\beta$ are decay constants equal to $\alpha \equiv exp(-\frac{\vartriangle_t}{\tau_{mem}})$ and $\beta \equiv exp(-\frac{\vartriangle_t}{\tau_{syn}})$ with a small simulation time step $\vartriangle_t >0$ and membrane and synaptic time constants $\tau_{mem}$ and $\tau_{syn}$; $W_{ij}$ are synaptic weights of the postsynaptic neuron $i$ and presynaptic neurons $j$; $V_{ij}$ are recurrent synaptic weights of neurons $i$ and $j$ within the same layer $l$; and $S_j^{(l)}[n]$ is the output spike train of neuron $j$ in layer $l$ at time step $[n]$. The output spike train is expressed as the Heaviside step function of the difference between the membrane voltage and the firing threshold $\varphi$ as follows: $\hat{Y}$

\begin{equation}\label{eq:OutputSpikeTrain}
S_j^{(l)}[n] = \Theta(U_j^{(l)}[n] - \varphi) 
\end{equation}

\subsection{Training}

The goal is to learn a set of weights, $W$, that predicts task 1 when using the firing threshold $\varphi = \varphi_1$ and task 2 (or, in general, more tasks) when $\varphi = \varphi_2$. In order to achieve this, during training, two random processes are used. First, a sample (or a sample batch) is randomly selected as the input. Second, a task to train for is randomly selected. If task 1 is selected, then the firing threshold of the feature extraction block and the label classifier block is set to $\varphi = \varphi_1$ and, similarly, for task 2 it is $\varphi = \varphi_2$. After setting the firing threshold, the input samples are presented. Backpropagation SLAYER \cite{Shrestha2018} algorithm is used to minimize both the label classifier and the task classifier loss functions. The total loss, $L$, is calculated as:

\begin{equation}\label{eq:lossEquation}
L = (1-\gamma)*L_y + \gamma*L_t 
\end{equation}

where $L_y$ is the loss for the label classifier block given by $L_y=Loss(Y, \hat{Y})$; $L_t$ is the loss for the task classifier block given by $L_t=Loss(T, \hat{T})$; and $\gamma$ is a loss rate constant that controls the rate between the label and task classifier losses. The true labels for the label classifier block, $Y$, are constructed as a concatenation of $Y_1$ and $Y_2=0$ or $Y_1=0$ and $Y_2$ depending on whether task 1 or task 2 was selected. The task classifier block is trained to predict a 0 or 1 depending on whether the network is being trained with task 1 or task 2 data. Note that we do not change the firing threshold of the task classifier block. This is because, we want the task classifier block to backpropagate the same information to the feature extraction block regardless of which task is being performed. 

\subsection{Testing}

For testing, first the firing threshold equal to $\varphi_1$ or $\varphi_2$ is set depending which task is tested. Second, the samples are input only to the feature extraction and to the label classifier blocks. The task classifier block is not used since the job is not to predict the task being performed as it is already determined by the chosen firing threshold.

\subsection{Implementation}

The network is implemented using Intel's framework Lava which consists of a set of libraries for development of neuromorphic simulation\footnote{The lava and lava-dl library are available at https://lava-nc.org}. Lava framework is designed to allow for deployment on the Loihi2 neuromorphic chip. The code for the network implementation as well as all experiments and results are posted at GitHub.\footnote{https://github.com/PaoloGCD/MultiTask-SNN}

\section{Experiments and results}

The performance of MT-SNN, which architecture is shown in Figure \ref{fig:MultitaskArchitecture}, is tested on the neuromorphic NMNIST data (60K training and 10K testing samples) \cite{Garrick2015}. Three types of experiments are performed. First, the training and testing performance of MT-SNN for different threshold values is tested (Fig. \ref{fig:TrainingThreshold} and Table \ref{tab:testingAccuracy}). The task classifier block is not used in order to asses the effects of the selected threshold values only. Second, the influence of including the task classifier block in training are analyzed (Table \ref{tab:3blockTesting}). Third, results of the MT-SNN and a MT-SNN that uses the external input current (not the threshold) are compared (Table \ref{tab:biasControl}).
    
\subsection{Varying threshold}

\begin{figure}[!ht]
\centering
\includegraphics[clip, width=0.9\columnwidth]{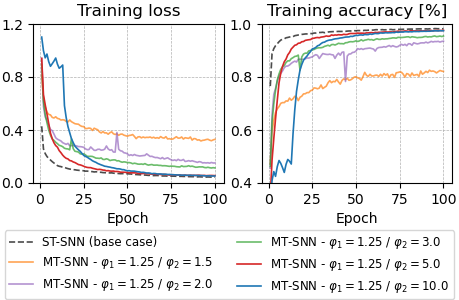}
\caption{Training accuracy and training loss for the ST-SNN (base case) and MT-SNN using different threshold values. $\varphi_1$ is kept at 1.25 and $\varphi_2$ is changed from 1.5 to 10.} 
\label{fig:TrainingThreshold}
\end{figure}

Figure \ref{fig:TrainingThreshold} shows the training loss and accuracy of the MT-SNN for two-task classification on the NMNIST data using different thresholds. For task 1, the digit classification with 10 labels, and for task 2 the odd/even digit classification with 2 labels are used. Figure \ref{fig:TrainingThreshold} also shows the results for a single-task spiking neural network, called ST-SNN, separately trained only on task 1 or task 2 as a base case. The network architecture for both MT-SNN and ST-SNN is essentially the same. It consists of two layers of 512 spiking neurons in the feature extraction block and two layers of 128 and 12 spiking neurons in the label classifier block. Note that $\varphi_1$ is kept at 1.25 in all tests while $\varphi_2$ varies from 1.5 to 10. The constant $\varphi_1$ value is used to tune spiking neurons to operate in a normal operation mode (single tasking).

Notice in Fig. \ref{fig:TrainingThreshold} that using $\varphi_1=1.25$ and $\varphi_2=5.0$ results in performance close to the base case scenario (when ST-SNN is trained on task 1 only). Using values for $\varphi_1$ and $\varphi_2$ close to each other achieves results in lower accuracies than the base case. On the other hand, using values that are too far apart (like $\varphi_1=1.25$ and $\varphi_2=10$) causes longer training time for MT-SNN. Table \ref{tab:testingAccuracy} compares accuracy on task 1 and task 2 of MT-SNN for different firing threshold pairs, after training for 100 epochs. Table \ref{tab:testingAccuracy} also includes accuracy for the ST-SNN (base case).

\begin{table}[ht!]
\centering
\caption{Testing accuracy on NMNIST data for MT-SNN using different firing threshold values.}
\resizebox{\columnwidth}{!}{%
\begin{tabular}{|l|cc|}
\hline
\multicolumn{1}{|c|}{\multirow{2}{*}{Model}} & \multicolumn{2}{c|}{Testing accuracy (\%)} \\ \cline{2-3} 
\multicolumn{1}{|c|}{}                     & \multicolumn{1}{c|}{Task 1} & Task 2 \\ \hline
ST-SNN (base case)                            & \multicolumn{1}{c|}{\textbf{98.85}} & \textbf{100.00} \\ \hline
MT-SNN - $\varphi_1=1.25$, $\varphi_2=1.5$  & \multicolumn{1}{c|}{93.73} & 98.00 \\ \hline
MT-SNN - $\varphi_1=1.25$, $\varphi_2=2.0$  & \multicolumn{1}{c|}{95.40} & 98.31 \\ \hline
MT-SNN - $\varphi_1=1.25$, $\varphi_2=3.0$  & \multicolumn{1}{c|}{96.60} & 98.90 \\ \hline
MT-SNN - $\varphi_1=1.25$, $\varphi_2=5.0$  & \multicolumn{1}{c|}{97.86} & \textbf{99.19} \\ \hline
MT-SNN - $\varphi_1=1.25$, $\varphi_2=10.0$ & \multicolumn{1}{c|}{\textbf{97.99}} & 99.13 \\ \hline
\end{tabular}%
}
\label{tab:testingAccuracy}
\end{table}

Two conclusions can be drawn from these results. First, similar to training performance shown in Figure \ref{fig:TrainingThreshold}, MT-SNN performs better when the difference between $\varphi_1$ and $\varphi_2$ increases. Second, the best accuracy on task 1 for MT-SNN, with $\varphi_1=1.25$ and $\varphi_2=10.0$, is $0.86\%$ lower than the accuracy of the base case. Such performance decrease is typical in multi-task problems. In order to improve this performance, we use the task classifier block, which results are presented next.

\subsection{Using the task classifier block in training}

The task classifier block is used  to improve the performance loss due to the nature of multi-task problems. Table \ref{tab:3blockTesting} compares accuracy with the use of task classifier block during training. Results are shown for the loss rate constant $\gamma$ (Equation \ref{eq:lossEquation}) values equal to 0.5, 0.3, 0.2 and 0.1.  For the convenience of the reader,  the results for ST-SNN and MT-SNN (repeated from  Table \ref{tab:testingAccuracy}) are also shown. All test are done using $\varphi_1=1.25$ and $\varphi_2=5$ values.

\begin{table}[ht!]
\centering
\caption{Testing accuracy on NMNIST data for MT-SNN using task classifier block}
\resizebox{\columnwidth}{!}{%
\begin{tabular}{|l|cc|}
\hline
\multicolumn{1}{|c|}{\multirow{2}{*}{Model}} & \multicolumn{2}{c|}{Testing accuracy (\%)}            \\ \cline{2-3} 
\multicolumn{1}{|c|}{}          & \multicolumn{1}{c|}{Task 1}         & Task 2          \\ \hline
ST-SNN - Base case                 & \multicolumn{1}{c|}{\textbf{98.85}} & \textbf{100.00} \\ \hline
MT-SNN (without task classifier) & \multicolumn{1}{c|}{97.86}          & 99.19           \\ \hline
MT-SNN / $\gamma = 0.1$                       & \multicolumn{1}{c|}{\textbf{97.97}} & \textbf{100.00} \\ \hline
MT-SNN / $\gamma = 0.2$          & \multicolumn{1}{c|}{97.72}          & 100.00          \\ \hline
MT-SNN / $\gamma = 0.3$          & \multicolumn{1}{c|}{97.59}          & 100.00          \\ \hline
MT-SNN / $\gamma = 0.5$          & \multicolumn{1}{c|}{97.69}          & 100.00          \\ \hline
\end{tabular}%
}
\label{tab:3blockTesting}
\end{table}

We see that the addition of the task classifier block increases performance of MT-SNN by $0.11\%$ on task 1 and by $0.81\%$ on task 2 for $\gamma=0.1$. The small increase can be attributed to the fact that the task classifier used in training is very simple (only two labels). It was observed during the experiments that the task classifier block reaches a plateau with $100\%$ accuracy after only 20 epochs; after that its contribution to the feature extraction block is minimal. 

\subsection{Using firing threshold vs external input current}

For another comparison Table \ref{tab:biasControl} shows accuracy of MT-SNN that uses the external input current (MT-SNN-EC) instead of the firing threshold for predicting task 1 or task 2. The training was done for 100 epochs using $I_{ext1}=0$ for task 1 and $I_{ext2}$ equal to 0.05, 0.1, 0.5, 1 and 5 for task 2.

\begin{table}[ht!]
\centering
\caption{Testing accuracy on NMNIST data for MT-SNN that uses $I_{ext}$ to control the network operation}
\resizebox{\columnwidth}{!}{%
\begin{tabular}{|l|cc|}
\hline
\multicolumn{1}{|c|}{\multirow{2}{*}{Model}} & \multicolumn{2}{c|}{Testing accuracy (\%)}            \\ \cline{2-3} 
\multicolumn{1}{|c|}{}         & \multicolumn{1}{c|}{Task 1}         & Task 2          \\ \hline
ST-SNN (base case)                & \multicolumn{1}{c|}{\textbf{98.85}} & \textbf{100.00} \\ \hline
MT-SNN / $\gamma = 0.1$                       & \multicolumn{1}{c|}{\textbf{97.97}} & \textbf{100.00} \\ \hline
MT-SNN-EC / $I_{ext2} = 0.05$ & \multicolumn{1}{c|}{95.63}          & 98.06           \\ \hline
MT-SNN-EC / $I_{ext2} = 0.1$  & \multicolumn{1}{c|}{96.05}          & 97.86           \\ \hline
MT-SNN-EC / $I_{ext2} = 0.5$                & \multicolumn{1}{c|}{\textbf{96.07}} & 97.66  \\ \hline
MT-SNN-EC / $I_{ext2} = 1.0$  & \multicolumn{1}{c|}{95.78}          & 97.62           \\ \hline
MT-SNN-EC / $I_{ext2} = 5.0$  & \multicolumn{1}{c|}{92.20}          & \textbf{97.95}           \\ \hline
\end{tabular}%
}
\label{tab:biasControl}
\end{table}

We notice that while controlling $I_{ext}$ the results are lower than when modifying the firing threshold of the spiking neurons. This behavior is interesting since it is closer to how biological neurons perform neuromodulation.

\section{Conclusion}

In this work we designed and implemented a multi-task spiking neuron network, MT-SNN, that uses the firing threshold to modify its operation while using the single tasking of multiple tasks approach. MT-SNN architecture consists of three processing blocks for feature extraction, for label classification and for task classification; the latter serving as a regularization process. Using the task classifier block in training resulted in small improvement of testing accuracy.  The results show that MT-CSNN  predicts both tasks with only slightly lower accuracy than ST-SNN. Specifically, MT-CSNN achieved 97.97\% accuracy while ST-CSNN 98.85\%. Experimental comparison of using the firing threshold vs using the external input current shows that with the firing threshold the accuracy is 97.97\% while with the external input current the accuracy is 96.07\%. Note that MT-CSNN was implemented on Intel’s Lava neuromorphic platform.

\bibliographystyle{IEEEtran.bst}
\bibliography{main}

\end{document}